\documentclass[10pt,twocolumn,letterpaper]{article}

\usepackage{cvpr}
\usepackage{times}
\usepackage{epsfig}
\usepackage{graphicx}
\usepackage{amsmath}
\usepackage{amssymb}

\usepackage[pagebackref=true,breaklinks=true,letterpaper=true,colorlinks,bookmarks=false]{hyperref}

\cvprfinalcopy 

\def\cvprPaperID{****} 

\ifcvprfinal\fi
\begin{document}

\title{Video Fill in the Blank with Merging LSTMs}

\author{Amir Mazaheri, Dong Zhang, Mubarak Shah\\
Center for Research in Computer Vision, University of Central Florida\\
Orlando, Florida, USA\\
{\tt\small mazaheri@cs.ucf.edu, dzhang@cs.ucf.edu, shah@crcv.ucf.edu}
}

\maketitle

\begin{abstract}
   Given a video and its incomplete textural description with missing words, the \textbf{Video-Fill-in-the-Blank (ViFitB)} task is to automatically find the missing word. The contextual information of the sentences are important to infer the missing words; the visual cues are even more crucial to get a more accurate inference. In this paper, we presents a new method which intuitively takes advantage of the structure of the sentences and employs merging LSTMs (to merge two LSTMs) to tackle the problem with embedded textural and visual cues. In the experiments, we have demonstrated the superior performance of the proposed method on the challenging ``Movie Fill-in-the-Blank'' dataset \cite{Rohrbach2016}.
\end{abstract}

\section{Introduction and Related Work}

\begin{figure}
\begin{center}
   \includegraphics[width=2.5in]{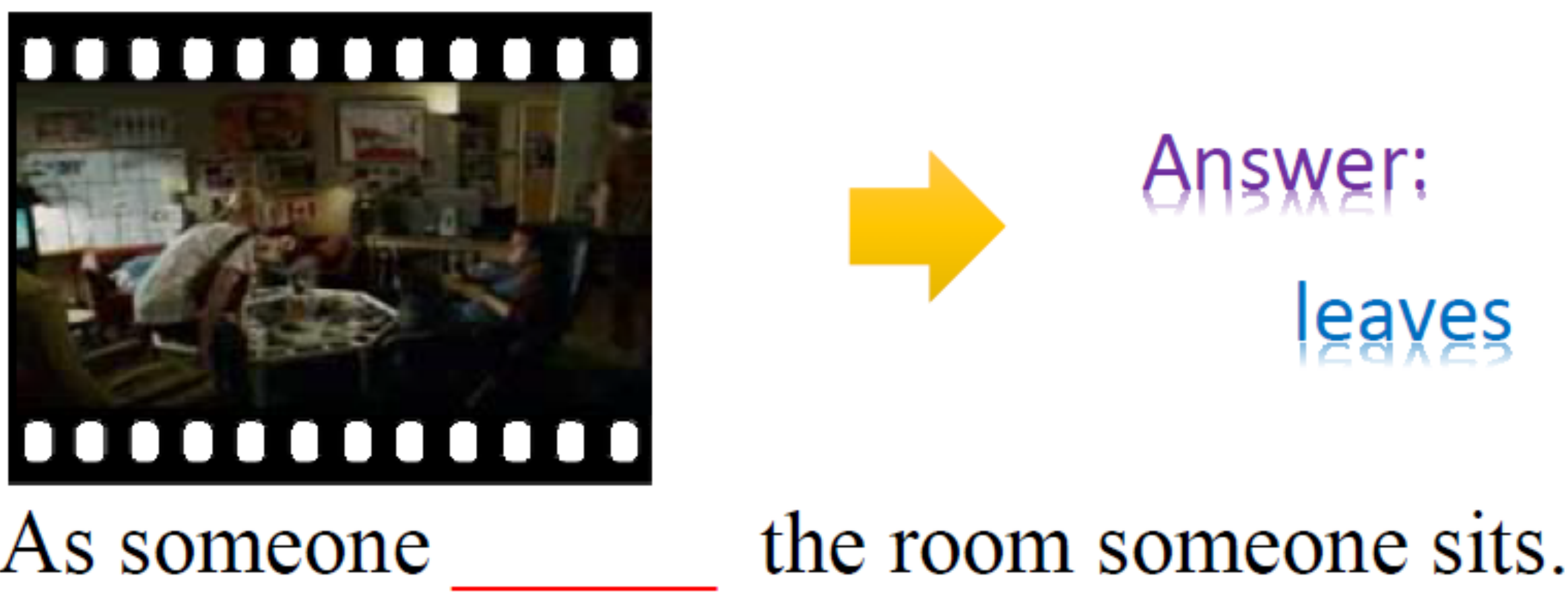}
\end{center}
   \caption{An example of the Video-Fill-in-the-Blank problem.}
\label{fig_example}
\end{figure}

Video-Fill-in-the-Blank (ViFitB) is a new computer vision problem. (Figure \ref{fig_example} shows an example). It is related to the standard Video-Question-and-Answer (ViQaA) problem, but have significant differences. A major difference is that, for a standard VQaA problem, it has a complete sentence of the ``question'', and requires the methods to output the ``answer''. In this framework, it is easier to encode the ``question'' in the model (e.g. Neural Networks) and then use it to predict an output. However, in the Video-Fill-in-the-Blank (ViFitB) problem, it is much trickier to encode the ``question'', since the ``question'' is broken into parts and is much more challenging to encode the pieces of question to find the missing word of ``blank'' efficiently.

In this paper, we propose a new method to encode the sentence fragments before and after the blank using two LSTMs (Long Short Term Memory), and use these two LSTMs' outputs to find the correct answer. This is an intuitive way to encode the semantic and contextual information from the sentence efficiently and makes the model easier to train. Similar to traditional Visual-Question-and-Answer (VQaA), we encode both textural and visual cues. From the experiments we found that both cues are crucial to improve the performance.

\begin{figure*}
\begin{center}
   \includegraphics[width=5in]{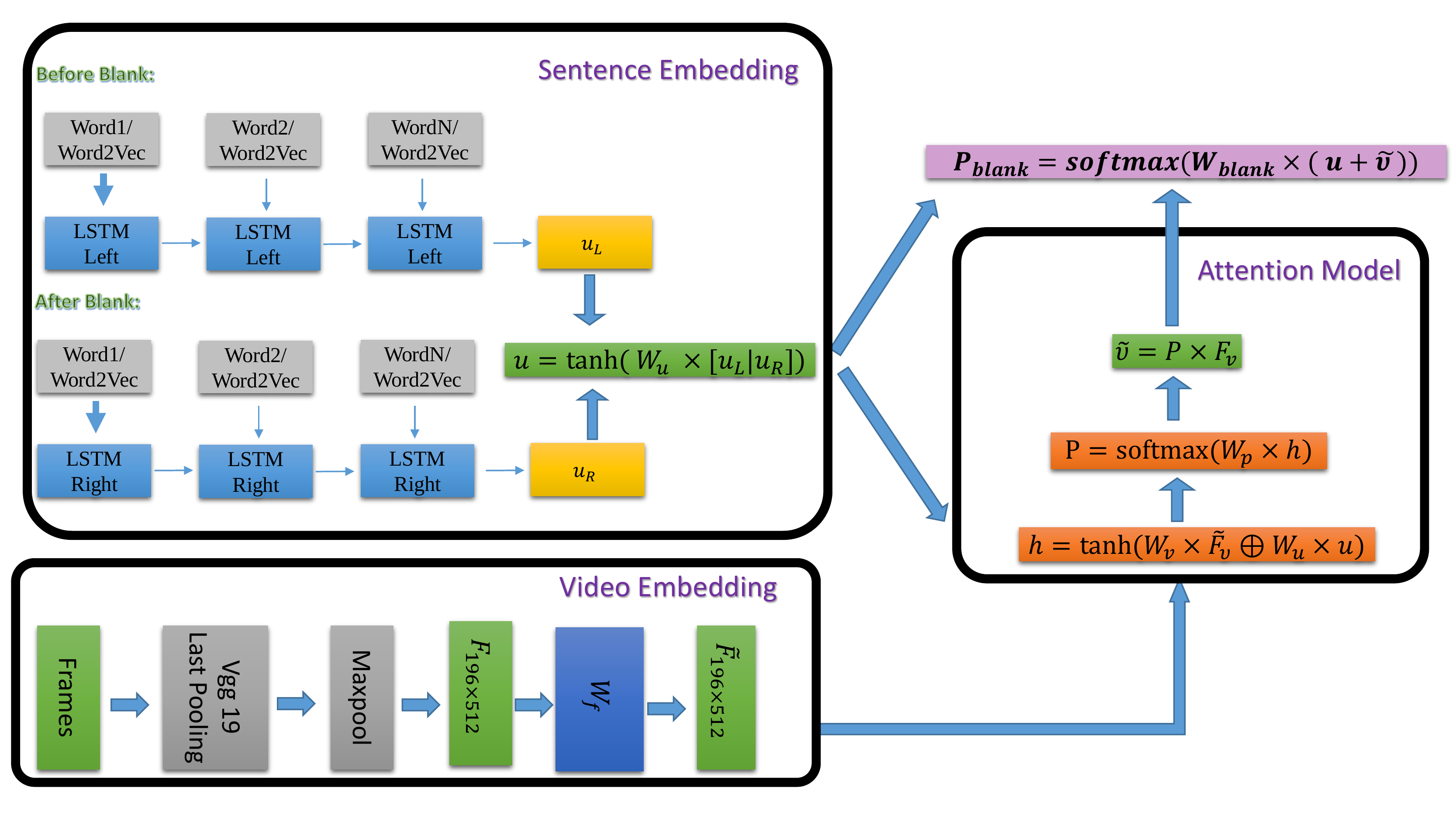}
\end{center}
   \caption{The Framework for the Proposed Method.}
\label{fig_framework}
\end{figure*}

The Visual-Fill-in-the-Blank (VFitB) problem is related to Visual-Captioning (VC) problem \cite{Xu2015,Venugopalan2014}, for which the deep neural networks, especially Recurrent Neural Networks (e.g. LSTM: Long Short Term Memory) are intensely employed \cite{Yao2015,Venugopalan2015}. It's also highly related to the Visual-Question-and-Answer (VQaA) problem \cite{Antol2015,Ren2015,Yang2015a,Kafle2016,Wu2015,Zhu2015}. For VQaA problem, usually there is a ``question'' with a question mark, and the answer is a word from the dictionary \cite{Antol2015}. The answer can be numbers, colors, objects, etc. Some of the datasets \cite{Zhu2015} provide multiple choices as the candidates for the answer. Strictly speaking, a Visual-Fill-in-the-Blank (VFitB) problem can be considered as a VQaA problem; however, this question is dramatically different from standard VQaA problems. There are already some work for Image-Fill-in-the-Blank (ImFitB) problem \cite{Yu2015}, however, the blanks in the dataset are usually at the end of the sentence and it is similar to sentence prediction. Also there is a dataset for Video-Fill-in-the-Blank (ViFitB) problem \cite{Zhu2015a}, however, the dataset was collected in a multiple choice fashion and limited number of words are available for the blank and mostly are about objects or actions. To our knowledge, ``Movie Fill-in-the-Blank'' dataset \cite{Rohrbach2016} is the first large-scale dataset for the ViFitB problem in the wild and we show superior performance of the proposed method on the dataset.

\section{Method}

The proposed method encodes the sentence with two LSTMs, and encodes the video by max-pooling over features coming out of a CNN. An attention model \cite{Yang2015} helps us to leverage more spatial information from videos. Figure \ref{fig_framework} shows an illustration of the proposed method. Each of the words from the sentence is converted into a feature vector using the Word2Vec approach \cite{Mikolov2013}, and then pass to LSTMs one by one. The left part (before blank) and the right part (after blank) are fed into two different LSTMs, aiming to capture the structure of the sentence. We use $u_L$ and $u_R$ to represent the output from the left and right LSTMs respectively, and we combine them using by
\begin{equation}
    u = tanh(W_u \times  [u_L | u_R]),
\end{equation}
where $u$ is the final output, $W_u$ is a trainable parameter matrix, and ``$|$'' means concatenation of left and right vectors. The output $u$ from the two LSTMs are then input into the visual attention model. On the other hand, the video frames are processed by a pre-trained VGG19 network \cite{Simonyan2014} and the last pooling layer is extracted. A max-pooling has been applied to all the frames' features and one $14 \times 14 \times 512$  tensor is the outcome of this part as a visual feature. Within the attention model, this visual feature and  the textural feature $u$ are combined by
\begin{equation}
    h = tanh(W_{vh} \times F_v \oplus W_{uh} \times u),
\end{equation}
where $h$ is output feature, $W_{vh}$ and $W_{uh}$ are trainable parameter matrices, $F_v$ is the visual feature, and ``$\oplus$'' means summation of the feature vector $u$ wit all $196$ vectors in $F_v$. The combined feature $h$ is then used to find attention model:
\begin{equation}
    P = softmax(W_p \times h),
\end{equation}
where $W_p$ is the trainable parameter matrix. The attention model $P$ is employed to output the weights for $196$ visual features. Finally, the textural and weighted visual features $\tilde{v} = P \times F_v$ are combined to predict the word in the blank
\begin{equation}
    P_{blank} = softmax(W_{blank} \times (u+\tilde{v})).
\end{equation}

\section{Experiments}

There are two methods to train the proposed deep network: End-to-End training, and Incremental training. For End-to-End training, we train the whole network together, and for Incremental training, we first train the sentence network (LSTMs), then combine it with the visual attention network and train them together again. We found the Incremental training method has better performance.

We did our experiments on the ``Movie Fill-in-the-Blank'' dataset \cite{Rohrbach2016}. We use Categorical Cross-Entropy as our loss function and ADaGrad optimizer with early stoping strategy by observing the validation loss.

We show the performances of the proposed method with different setups, and compared it with several state-of-the-art methods. Table \ref{table_results} shows some of the quantitative results.
The ``End-to-End" is our method which trains all the parameters from scratch. However, Our ``Incremental" method uses the method ``Sentence" to initialize two LSTMs and the matrix $W_u$, since they are in common between both of them. The ``Left Sentence" method just uses the left part of the blank to answer the question. ``Visual LSTM" method uses just CNN features coming out of last fully connected layer of Vgg-19 and pass them through an LSTM to get the answer with out considering the question.

\begin{table}
\begin{center}
\begin{tabular}{ll|c}
\hline
Method & Setup & Accuracy \\
\hline\hline
LSTM & Sentence      & 0.280 \\
LSTM & Visual        & 0.055 \\
LSTM & Left Sentence & 0.155 \\
LSTM & Video+Sentences & 0.312\\
Ours & End-to-End    & 0.317 \\
Ours & Incremental   & 0.342 \\

\hline
\end{tabular}
\end{center}
\label{table_results}
\caption{Results on ``Movie Fill-in-the-Blank'' dataset.}
\end{table}

\section{Conclusion}

We have proposed a new method for the Video-Fill-in-the-Blank (ViFitB) problem which takes advantage of the sentence structure before and after the blank and employed two LSTMs to encode the textural information efficiently. We have demonstrated that, by incorporating the visual cues with the spatial attention model, the performance can be further improved. We verified our ideas on the new ``Movie Fill-in-the-Blank'' dataset of 2016' Large Scale Movie Description and Understanding Challenge (LSMDC), and showed improved results compared with several baseline methods.
In experiments, we have considered the output dimension of $u$ vector as $1000$. Also the batch size in training stage is $16$. Each epoch takes about $2000$ seconds in our implementation settings.


\end{document}